\pdfoutput=1

\documentclass[11pt]{article}

\usepackage{acl}
\usepackage{multirow}
\usepackage{amsmath}
\usepackage{fontenc}

\usepackage{times}
\usepackage{latexsym}
\usepackage{multirow}
\usepackage{todonotes}
\usepackage{enumitem}

\usepackage{subcaption}

\usepackage[T1]{fontenc}

\usepackage[utf8]{inputenc}

\usepackage{microtype}

\usepackage{inconsolata}

\usepackage{graphicx}

\usepackage{booktabs}
\usepackage{diagbox}
\usepackage{CJK}
\usepackage{xspace}
\usepackage{enumitem}
\usepackage{amssymb}
\usepackage{cleveref}
\usepackage{listings}
\usepackage{xcolor}
\usepackage{adjustbox}

\definecolor{notebookbg}{RGB}{248,248,248}

\usepackage{xcolor}        
\usepackage{listings}
\definecolor{mdback}{RGB}{250,248,240}   
\definecolor{mdtext}{RGB}{ 50, 50, 50}   
\definecolor{mdhead}{RGB}{  0, 92,197}   
\definecolor{mdquote}{RGB}{ 98,116, 85}  

\lstdefinelanguage{markdown}{
  morecomment = [l][\color{mdhead}\bfseries]{\#},   
  morecomment = [l][\color{mdhead}]{>},             
  morecomment = [l][\color{mdhead}]{-},             
  morecomment = [l][\color{mdhead}]{*},             
  morestring  = [b][\color{mdcode}]{`},
  alsoletter  = {-},
  sensitive   = false
}

\lstdefinestyle{mdstyle}{
  language        = Markdown,
  basicstyle      = \ttfamily\small\color{mdtext},
  backgroundcolor = \color{mdback},
  frame           = single,
  rulecolor       = \color{black!15},
  breaklines      = true,
  showstringspaces= false,
  numbers         = none,
  morecomment     = [l][\color{mdquote}]{>},      
  morecomment     = [l][\color{mdhead}]{\#},      
  morecomment     = [l][\color{mdhead}]{-},       
}

\newcommand{\multirowcell}[1]{\begin{tabular}[c]{@{}c@{}}#1\end{tabular}}
\newcommand{\ex}[1]{\textit{#1}\xspace}

\title{A Head to Predict and a Head to Question:
\\Pre-trained Uncertainty Quantification Heads for \\ Hallucination Detection in LLM Outputs}

\author{
Artem Shelmanov\textsuperscript{1} \quad Ekaterina Fadeeva\textsuperscript{2} \quad Akim Tsvigun\textsuperscript{4} \quad Ivan Tsvigun\textsuperscript{5}, \\
\textbf{Zhuohan Xie\textsuperscript{1} \quad Igor Kiselev\textsuperscript{6} \quad Nico Daheim\textsuperscript{2} \quad Caiqi Zhang\textsuperscript{3} \quad Artem Vazhentsev\textsuperscript{7}} \\ \textbf{Mrinmaya Sachan\textsuperscript{2} \quad Preslav Nakov\textsuperscript{1} \quad Timothy Baldwin\textsuperscript{1}}\\
\textsuperscript{1}MBZUAI \quad \textsuperscript{2}ETH Zürich \quad  \textsuperscript{3}University of Cambridge \\ \textsuperscript{4}Nebius.AI \quad \textsuperscript{5}Behavox \quad  \textsuperscript{6}Accenture\\
\textsuperscript{7}Computational Semantics Group
\\
\href{mailto:artem.shelmanov@mbzuai.ac.ae}{\{artem.shelmanov,preslav.nakov,timothy.baldwin\}@mbzuai.ac.ae} \\
\href{mailto:efadeeva@ethz.ch}{\{efadeeva,msachan\}@ethz.ch} \quad \href{mailto:aktsvigun@nebius.com}{aktsvigun@nebius.com}
}

\begin{document}
\maketitle
\begin{abstract}
Large Language Models (LLMs) have the tendency to hallucinate, i.e., to sporadically generate false or fabricated information. This presents a major challenge, as hallucinations often appear highly convincing and users generally lack the tools to detect them. Uncertainty quantification (UQ) provides a framework for assessing the reliability of model outputs, aiding in the identification of potential hallucinations. In this work, we introduce pre-trained UQ heads: supervised auxiliary modules for LLMs that substantially enhance their ability to capture uncertainty compared to unsupervised UQ methods. Their strong performance stems from the powerful Transformer architecture in their design and informative features derived from LLM attention maps. Experimental evaluation shows that these heads are highly robust and achieve state-of-the-art performance in claim-level hallucination detection across both in-domain and out-of-domain prompts. Moreover, these modules demonstrate strong generalization to languages they were not explicitly trained on.
We pre-train a collection of UQ heads for popular LLM series, including Mistral, Llama, and Gemma 2. We publicly release both the code and the pre-trained heads.\footnote{\url{https://github.com/IINemo/llm-uncertainty-head}}
\end{abstract}

\begin{figure*}[t!]
    \centering
    \resizebox{0.75\textwidth}{!}{
    \includegraphics[trim={0.cm 0.cm 1.6cm 0.cm},clip,width=1.\linewidth]{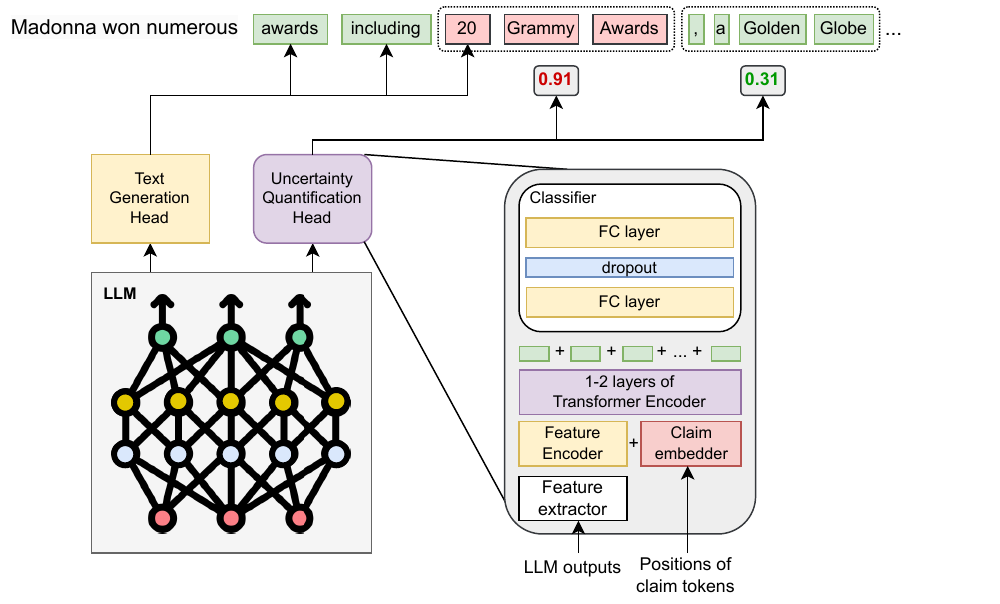}}
    \caption{The architecture of uncertainty quantification heads. The example represents a text generated using an LLM, containing the hallucination \ex{20 Grammy Awards} highlighted in red.}
    \label{fig:uncertainty_head}
\end{figure*}

\section{Introduction}

Uncertainty quantification (UQ)~\cite{gal2016dropout,baan2023uncertainty,geng-etal-2024-survey,
zhang-etal-2024-luq}
has become an increasingly important topic in natural language processing (NLP), particularly for addressing
challenges with hallucinations \cite{huang2025survey} and low-quality outputs
of large language models (LLMs)~\cite{malinin2020uncertainty,kuhn2023semantic,fadeeva2024fact}. UQ offers the potential to improve the safety and reliability of LLM-based
applications by flagging highly uncertain generations. Such generations could be discarded
or marked as untrustworthy, thus reducing the risk of  misleading information reaching users \citep{zhang-etal-2024-luq, zhang2024atomic, huang-etal-2024-calibrating}.
Contrary to other methods for detecting hallucinations that rely on external knowledge bases or additional LLMs~\cite{manakul-etal-2023-selfcheckgpt,min2023factscore,chen2023hallucination}, UQ methods assume that LLMs naturally encode information about their own limitations, and this self-knowledge can be efficiently accessed.

There are many existing UQ techniques for well-defined tasks such as classification and regression
~\cite{zhang-etal-2019-mitigating,he2020towards,xin-etal-2021-art,wang-etal-2022-uncertainty,vazhentsev-etal-2023-hybrid,he-etal-2024-uncertainty}.
However, applying UQ to text generation has unique challenges including (\emph{i})~an infinite
number of possible generations, which complicates the normalization of the uncertainty scores, (\emph{ii})~potentially multiple correct answers with different surface forms \cite{kuhn2023semantic},  (\emph{iii})~need to aggregate uncertainties across multiple interdependent predictions corresponding to generated tokens~\cite{zhang2023enhancing}, (\emph{iv})~generated tokens not contributing to uncertainty equally, as some tokens represent auxiliary words \cite{duan-etal-2024-shifting}, and (\emph{v})~some sources of uncertainty being irrelevant for hallucination detection \cite{fadeeva2024fact}.
These challenges hinder the performance of classical unsupervised UQ techniques, and addressing them explicitly in a single method is quite difficult.
Recently, researchers have proposed learning the aforementioned intricacies from the annotated data and developed supervised
methods for UQ and hallucination detection \cite{azaria-mitchell-2023-internal,li-etal-2024-reference,he2024llm,chuang2024lookbacklensdetectingmitigating}.

We continue this line of work by introducing pre-trained UQ heads: supervised auxiliary modules for LLMs that substantially enhance their ability to capture uncertainty compared to unsupervised UQ methods.
Their strong performance stems from the powerful Transformer architecture in their design and informative features derived from LLM attention maps. These heads do not require re-training of the entire LLM and do not alter its outputs. In addition to their high performance, these methods maintain a relatively small memory and computational footprint, ensuring practical usability.

Experimental evaluation shows that our uncertainty heads are highly robust and achieve state-of-the-art performance in claim-level hallucination detection across both in-domain and out-of-domain prompts, outperforming other supervised and unsupervised techniques. Moreover, these modules demonstrate strong generalization to languages they were not explicitly trained on.

Training uncertainty quantification heads requires annotated hallucinations in LLM outputs. For constructing training data, we created an automatic pipeline for the annotation of hallucinations of LLM outputs, which allows us to scale our experiments and to pre-train uncertainty heads for various LLMs. We release a collection of pre-trained UQ heads for popular open-source instruction-following LLMs, including Llama series \cite{dubey2024llama}, Gemma 2 \cite{team2023gemini}, and Mistral-v0.2 \cite{jiang2023mistral}.

The contributions of this work are as follows:
\begin{itemize}[topsep=0pt,itemsep=-1ex,partopsep=1ex,parsep=1ex]
    \item We design a pre-trained uncertainty quantification head: a supplementary module for an LLM that yields substantially better performance for claim-level hallucination detection than classical unsupervised UQ methods and state-of-the-art supervised techniques.
    \item We conduct a vast empirical investigation and find that uncertainty heads show good generalization across various domains and languages. We also compare various feature sets used for building supervised UQ modules.
    \item We build and release a collection of pre-trained uncertainty quantification heads for popular series of open-source instruction-tuned LLMs: Llama, Gemma~2, Mistral. These modules could be seamlessly integrated into text generation code and be used as off-the-shelf hallucination detection tools.
\end{itemize}

\section{Related Work}

\paragraph{Unsupervised methods.}

UQ for LLMs has recently
experienced a surge of work, with early efforts focusing on unsupervised techniques such as information-based approaches \cite{kuhn2023semantic,farquhar2024detecting}, density-based scores \cite{vazhentsev-etal-2022-uncertainty}, self-consistency methods \cite{lin2023generating, zhang-etal-2024-luq, NEURIPS2024_f26d4fba}, and verbalized (reflexive) strategies \cite{tian-etal-2023-just}.
While unsupervised approaches have shown some potential, they still fall short of offering a strong solution to the problem of LLM hallucinations \cite{vashurin2024benchmarking}.

\paragraph{Supervised methods.}

Recently, researchers have started exploring supervised UQ methods that leverage the internal states of LLMs during generation as features \cite{azaria-mitchell-2023-internal,slobodkin-etal-2023-curious,su-etal-2024-unsupervised,ch-wang-etal-2024-androids,he2024llm}. These methods achieve substantial performance gains over unsupervised approaches, especially for in-domain data.

\citet{azaria-mitchell-2023-internal} proposed one of the first methods of this kind called SAPLMA, where they trained a perceptron using activations from various layers to
detect when the LLM ``agrees'' with false statements. \citet{slobodkin-etal-2023-curious} trained a linear model on hidden states to detect question ``answerability'', effectively identifying unanswerable questions that typically lead to hallucinations.

Factoscope \cite{he2024llm} implemented a Siamese model with a rich feature set that incorporates activation maps, token ranks, and probabilities from unembedding matrices across layers.
They reported performance improvements over SAPLMA within the training domain, but encountered challenges with generalization to other domains.

\citet{ch-wang-etal-2024-androids} trained simple linear and attention-based models (probes) for span-level hallucination detection, using manually annotated responses from multiple LLMs. They also tried to use synthetically generated data but found the results to be inferior to manual annotation, which limits the applicability of their approach.

Lookbacklens \cite{chuang2024lookbacklensdetectingmitigating} introduces a feature set derived from LLM attention maps. They calculate the ratio of attention weights for newly generated tokens to those in the input prompt. The ratios, computed across all attention heads and layers, are used in a linear regression model to predict an uncertainty score.

\paragraph{Limitations of Previous Methods.} While all these works introduced a number of valuable ideas, they have notable limitations. \citet{azaria-mitchell-2023-internal,slobodkin-etal-2023-curious,su-etal-2024-unsupervised} focused on sequence-level methods and are not able to detect sub-sentence hallucinations. Many models, including \citet{slobodkin-etal-2023-curious,azaria-mitchell-2023-internal,chuang2024lookbacklensdetectingmitigating,su-etal-2024-unsupervised} used non-contextualized architectures such as simple linear probes or multi-layer perceptron. Although \citet{he2024llm} integrated a linear model with an attention mechanism and \citet{ch-wang-etal-2024-androids} used a contextualized model combining convolutions, ResNet, and GRU, these architectures are considered outdated and exhibit limitations in quality or computational efficiency. The features of the majority of models included only hidden states across layers \cite{azaria-mitchell-2023-internal,slobodkin-etal-2023-curious,ch-wang-etal-2024-androids,su-etal-2024-unsupervised}, which limits their generalization. Only
\citet{he2024llm} and \citet{chuang2024lookbacklensdetectingmitigating} performed more elaborate feature engineering. Finally, synthetic data that is leveraged through enforced decoding is used in some work \cite{azaria-mitchell-2023-internal,slobodkin-etal-2023-curious}. Compared to the native outputs generated by LLMs, such data may introduce additional biases and adversely affect the performance of hallucination detectors.

In contrast, here we aim to build uncertainty quantification heads for subsentence hallucination detection: on the level of atomic claims that leverage all the strengths of the aforementioned work and address their limitations: (\emph{i})~instead of over-simple or outdated architectures, we build our solution on the powerful Transformer architecture, (\emph{ii})~we investigate the importance of various features for hallucination detection, finding that the most informative features are derived from attention maps of LLMs, and (\emph{iii})~we build an automatic pipeline for generating training data using the native LLM responses. This pipeline allows us to build training data at a larger scale and pre-train UQ heads for a range of popular LLMs.

\section{Uncertainty Quantification Head}

Consider the LLM $P(t_i \mid x,t_{<i})$ with $L$ layers receiving a prompt $x$ of length $n$ and generating tokens $y=\{t_1, t_2, ..., t_T\}$. We also have a set of atomic claims $C=\{c_1, c_2, ..., c_K\}$, each representing a mapping to a subset of tokens in the output. Atomic claims, for example, can be extracted by another light-weight model. In this work, we formalize the claim-level uncertainty quantification task as building a function $U(c_i | x, y) \in [0, 1]$ that determines whether the claim  $c_i \in C$  is a hallucination. A large value of $U(c_i|x,y)$ indicates a higher likelihood that the claim \( c_i \) is a hallucination.

\subsection{Background on Features for UQ and Hallucination Detection}

\paragraph{Hidden states} have been shown to serve as indicators of hallucinations in several studies \cite{azaria-mitchell-2023-internal,ch-wang-etal-2024-androids}. Hidden states $h(t)$ could be extracted from multiple layers of the LLM and aggregated, e.g., as a concatenation in a feature vector:
\begin{equation}
F_\text{hs}(t)=h_1(t) \circ h_1(t) \circ ... \circ h_L(t).
\end{equation}

\paragraph{Lookbacklens (LBLens).} \citet{chuang2024lookbacklensdetectingmitigating} leverage features derived from the LLM's attention maps. The key idea is that when the model attends to the prompt, it attempts to solve the task, whereas attending to generated tokens causes it to disregard the prompt, increasing the likelihood of hallucination. The authors suggest using the so-called lookback ratio -- the ratio of aggregated attention to tokens of the prompt and the generated tokens. Consider each layer of the LLM contains $Q$ attention heads, and $q$ is an index of a head.
$A^{q, l}_{\text{context}}(t_i)$ and $A^{q, l}_{\text{gen}}(t_i)$ are the average attention weights to the input $x$ and to the previously generated output $t_{<i}$, respectively:
$$
A^{q, l}_{\text{context}}(t_i) = \frac{1}{n} \sum_{j=1}^n \alpha^{q, l}_{t_i, x_j},
$$
$$
A^{q, l}_{\text{gen}}(t_i) = \frac{1}{i - 1} \sum_{j=n+1}^{i-1} \alpha^{q, l}_{t_i, t_j}.
$$
Here, $\alpha^{h, l}_{t_i, t_j}$ represents the softmax-weighted attention score from token $t_i$ to token $t_j$.

Then the lookback ratio of the model head \(q\) and the layer \(l\) for  the token \(t_i\) is defined as follows:
$$
    LR^{q, l}(t_i) = \frac{A^{q, l}_{\text{context}}(t_i)}{A^{q, l}_{\text{context}}(t_i) + A^{q, l}_{\text{gen}}(t_i)},
$$
\begin{equation}
F_\text{LBLens}(t_i)=\{LR^{q,l}(t_i)\}_{q,l}^{Q,L}.
\end{equation}

\paragraph{Factoscope.}
\citet{min2023factscore}, in addition to model activations, introduced a set of features that leverage token probabilities, the similarity of token embeddings across layers, and the evolution of token ranks across layers. Commonly, given a token $t_i$ at the position $i$, the LLM outputs hidden states $\{h_l(t_i)\}_{l=1}^L$, where the final hidden state $h_L(t_i)$ is passed through the unembedding matrix $E$ to predict token logits. Factoscope applies $E$ to each LLM layer, obtaining a set of token logits on a specific layer $l$:
$
z^l_i = E\left(h_l(t_i)\right).
$
Then, it extracts the logits of the top-$m$ tokens from each layer $l$:
\begin{equation}
    F_\text{top-tokens}(t_i) = \left\{z^l_i(t) \mid  t \in \text{top}_m(z^l_i)\right\}_{l=1}^L.
\end{equation}

To analyze token evolution across layers, Factoscope computes the cosine similarities between embeddings of top tokens from adjacent layers obtained by applying the unembedding matrix:
\begin{equation*}
\begin{aligned}
S^l(t_i) = \{ & \cos(E_{w_1}, E_{w_2}) \mid \\
& w_1 \in \text{top}_m(z^l_i), w_2 \in \text{top}_m(z^{l+1}_i) \}
\end{aligned}
\end{equation*}
\begin{equation}
    F_\text{tokens-sim}(t_i) = \{S^l(t_i)\}_{l=1}^{L-1}.
\end{equation}

Finally, Factoscope tracks token rank evolution across layers:
$
R^l(t_i) = \text{rank}[t_i, z_i^l],
$
where $\text{rank}$ indicates the position of $t_i$ in the descending order of $z_i^l$ values (top-ranked token receives 1). The ranks are further normalized to the range $[0,1]$:
\begin{equation}
    F_\text{rank}(t_i) = \{R^l(t_i)^{-1}\}_{l=1}^L.
\end{equation}

\subsection{Features for Pre-trained Uncertainty Quantification Heads}

We experimented with all the aforementioned types of features and their combinations. However, we found that all of them exhibited various limitations. Hidden states encode a lot of domain-specific information, increasing the risk of overfitting. Factoscope features usually require substantial computational overhead and do not add much new information compared to hidden states. Attention features are quite powerful, but aggregation suggested in Lookback lens results in the loss of valuable information.
For our pre-trained uncertainty quantification heads, we use two groups of features.

\paragraph{Attention maps of the LLM.}
Attention seems to carry the key information about LLM uncertainty,  which might be due to various reasons, including the fact that attention weights reflect the conditional dependency between the generation steps.
For each token, we obtain the attention maps to $k$ previous tokens from each attention head and layer and flatten them into a single feature vector:
\begin{equation}
    F_{\text{att}}=\{\alpha^{q, l}_{t_i, t_{i-j}}\}_{i,j,q,l}^{n,k,Q,L}.
\end{equation}
When $(i-j)$ is negative, we pad the feature vector with a zero placeholder.
While considering many previous tokens that might explode the size of the feature space, we empirically found that the optimal value of $k$ is typically very small: $2\le k \le 5$ (see Figure \ref{fig:an1}). We believe that this is due to the powerful contextualized architecture of our model that leverages a transformer to automatically extract useful patterns across the whole generated sequence.

\paragraph{Probability distributions of the LLM.} Although the probability distribution of an LLM can be misleading, it still conveys useful information about the model’s conditional confidence at the current generation step. This group of features consists of logarithms of the top-$m$ token probabilities:
\begin{equation}
\begin{aligned}
F_{\text{prob}}(t_i) = \{ & \log P(t \mid x, t_{<i}) \mid \\
& t \in \text{top}_m(P(\cdot \mid x, t_{<i}))\}.\\
\end{aligned}
\end{equation}
We concatenate all groups of features into a token-level feature vector: $F(t)=F_{\text{att}}(t) \circ F_{\text{prob}}(t)$.

\subsection{Architecture of Uncertainty Quantification Heads}

The architecture of the UQ head is depicted in \Cref{fig:uncertainty_head}. To ensure flexibility and expressive capacity, we build it on top of a Transformer backbone.
It consists of a feature size reduction network,
a multi-layer transformer encoder, and a two-layer classification neural network. For each component, we use GELU activation functions and dropout regularization. To mark tokens as belonging to the claim being classified, we introduce an embedding matrix. Each token, depending on whether it belongs to the classified claim, receives a corresponding embedding that is summed up with the representation from the feature size reduction network. The resulting representations are fed into the transformer encoder. The outputs of the encoder are averaged across all tokens of the claim and fed into the classifier. The UQ head is trained using a binary cross-entropy loss function. When we train heads, we freeze the ``body'' of the LLM, so that the LLM generations stay exactly the same.

\section{Pipeline for Training Data Generation}

The training data generation pipeline is presented in \Cref{fig:data_generation_pipeline} in the appendix.
It starts with prompting the LLM to produce responses for a list of questions such as \ex{Write a biography of person X} or \ex{Write the history of the city Y}. We select relatively famous named entities so the task is not very hard for the model based on its parametric knowledge, while at the same time, it is not trivial, so outputs contain some hallucinated claims.
We also do not use synthetically-generated hallucinations, as they introduce a bias between what the model actually generates vs. the synthetic data. The prompts for other domains can be found in \Cref{tab:test_stats}.

We split the obtained responses into atomic claims using GPT-4o with the prompts from \cite{fadeeva2024fact,vashurin2024benchmarking}. Each claim is then automatically classified by GPT-4o as \emph{supported}, \emph{unsupported}, or \emph{unknown}. The last category is intended for general claims, for which estimating the veracity is meaningless. The claim labeling process is two-staged: in the first stage, we ask the model to provide an elaborated answer via chain-of-thought (CoT), and in the second stage, we ask it to summarize its answer into one word. The performance of this two-stage labeling is substantially better than for one-stage labeling, due to the well-known issue of lack of logical reasoning in LLMs without CoT \cite{wei2022chain}.

The pipeline allows constructing relatively large-scale datasets annotated with claim-level hallucinations for various LLMs.
The budget for annotating the responses of one LLM for the biggest biographies dataset with 3,300 prompts was around \$100. While we used GPT-4o due to its strong performance for the annotation of hallucinations in prior studies \cite{vashurin2024benchmarking}, annotation quality could be further improved by leveraging more powerful LLMs or employing an ensemble of models.
Statistics about the training data used in our experiments are presented in \Cref{tab:train_stats}.

\begin{table*}[t]

\centering
\footnotesize

\resizebox{1\textwidth}{!}{

\begin{tabular}{l|cccccccc}
\toprule
\diagbox{\textbf{Method}}{\textbf{Test Sets}} & \multirowcell{\textbf{Biographies} \\ \textbf{(in domain)}} & \textbf{Cities} & \textbf{Movies} & \textbf{Inventions} & \textbf{Books} & \textbf{Artworks} & \textbf{Landmarks} & \textbf{Events} \\
\midrule
Random & 0.29 & 0.21 & 0.10 & 0.16 & 0.11 & 0.26 & 0.12 & 0.11 \\
MCP & 0.41 & 0.31 & 0.20 & 0.32 & 0.14 & 0.32 & 0.14 & 0.14 \\
Perplexity & 0.36 & 0.23 & 0.17 & 0.23 & 0.14 & 0.34 & 0.13 & 0.12 \\
Mean Token Entropy & 0.42 & 0.29 & 0.24 & 0.38 & 0.17 & 0.32 & 0.14 & 0.16 \\
CCP & 0.50 & 0.37 & 0.27 & 0.38 & 0.17 & 0.38 & 0.20 & 0.17 \\
\midrule
SAPLMA & 0.54 & 0.43 & 0.27 & 0.35 & 0.29 & 0.53 & \textbf{0.35} & 0.24 \\
Factoscope & \underline{0.61} & \underline{0.47} & \underline{0.34} & \underline{0.42} & \underline{0.32} & \underline{0.49} & 0.28 & 0.26 \\
Lookback lens & 0.56 & 0.45 & 0.25 & 0.39 & 0.26 & 0.46 & 0.26 & \underline{0.29} \\
UHead (Ours) & \textbf{0.66} & \textbf{0.49} & \textbf{0.47} & \textbf{0.48} & \textbf{0.40} & \textbf{0.56} & \underline{0.34} & \textbf{0.37} \\
\bottomrule
\end{tabular}

}

\caption{PR-AUC for various UQ methods for hallucination detection of the Mistral 7B Instruct v0.2 model on English datasets. Biographies represent the in-domain dataset for supervised UQ methods. 
}
\label{tab:results_train_bio_mistral}
\end{table*}
\begin{table*}[t]

\centering
\footnotesize

\resizebox{0.88\textwidth}{!}{

\begin{tabular}{l|ccccccc}
\toprule
\diagbox{\textbf{Method}}{\textbf{Test Sets}}  & \textbf{Cities} & \textbf{Movies} & \textbf{Inventions} & \textbf{Books} & \textbf{Artworks} & \textbf{Landmarks} & \textbf{Events} \\
\midrule
UHead, bio (new) & 0.49 & 0.47 & 0.48 & 0.40 & 0.56 & 0.34 & 0.37 \\
UHead, bio + all - 1 (new) & 0.49 & 0.48 & 0.48 & 0.40 & 0.57 & 0.34 & 0.39 \\
\bottomrule
\end{tabular}

}

\caption{Introducing more diverse training data. UHead results are shown for two scenarios: when the UQ head is trained solely on the English biographies dataset, and when it is trained on the biographies dataset along with all other domains, excluding the test domain.}
\label{tab:results_moredata}
\end{table*}
\begin{table}[t]

\centering
\footnotesize


\begin{tabular}{l|cc}
\toprule
\diagbox{\textbf{Method}}{\textbf{Test Set}} & \textbf{Biographies (dev)} \\
\midrule



UHead (only hidden states) & 58.2 \\
UHead (att. + probs. + hs.) & 58.9 \\
UHead (Factoscope) & 58.8 \\
UHead (LookBack Lens) & 60.9 \\
UHead (att. + probs.) & \textbf{64.2} \\

\bottomrule
\end{tabular}


\caption{PR-AUC scores for UQ heads trained with various feature sets on the Mistral 7B Instruct v0.2 model. Performance was evaluated using the validation set of the Biographies domain after hyperparameter tuning. }
\label{tab:results_train_bio_features}
\end{table}
\begin{table}[t]

\centering
\footnotesize

\resizebox{0.49\textwidth}{!}{

\begin{tabular}{l|cccc}
\toprule
\diagbox{\textbf{Method}}{\textbf{Language}} & 
\multirowcell{\textbf{English} \\ \textbf{(in domain)}}
& \textbf{Russian} & \textbf{Chinese} & \textbf{German} \\ 
\midrule

Random & 0.13 & 0.34 & 0.23 & 0.15 \\ 
MCP & 0.18 & 0.43 & 0.31 & 0.20 \\ 
Perplexity & 0.14 & 0.40 & 0.29 & 0.15 \\ 
Mean Token Entropy & 0.20 & 0.44 & 0.44 & 0.22 \\ 
CCP & 0.31 & 0.49 & 0.44 & 0.31 \\ 
\midrule
SAPLMA & 0.34 & 0.51 & 0.33 & \underline{0.39} \\
Factoscope & 0.35 & 0.53 & 0.35 & 0.38 \\
Lookback lens & \underline{0.36} & \textbf{0.58} & \underline{0.48} & \underline{0.39} \\
UHead (Ours) & \textbf{0.46} & \textbf{0.58} & \textbf{0.54} & \textbf{0.44} \\

\bottomrule
\end{tabular}

}

\caption{Performance comparison of the UQ head on different languages using the Gemma 2 9b Instruct model trained on English-only biographies data.}
\label{tab:results_multilang}
\end{table}

\section{Experiments}

\subsection{Experimental Setup}

For experiments, we used the LM-Polygraph framework
\cite{fadeeva2023lm}, which makes it easy to evaluate UQ for LLMs in a consistent way.

\paragraph{Evaluation Datasets.}

We constructed eight test sets of English questions designed to prompt LLMs to generate texts across various domains: \emph{person biographies}, \emph{cities}, \emph{movies}, \emph{inventions}, \emph{books}, \emph{artworks}, \emph{landmarks}, and \emph{events}.
Each test set contains 100 questions, generated by prompting GPT-4o and Claude-3-Opus to output 100 famous domain items, e.g., 100 famous landmarks. An example structure of the prompts we used is presented in \Cref{sec:app-dataset-constr}.\footnote{All data used for training and testing is available at \url{https://huggingface.co/llm-uncertainty-head}}
The labels for the test sets are obtained in the same way as the training sets: we generate responses using the LLM, automatically split the responses into atomic claims, and label them using GPT-4o.

To assess the cross-lingual generalizability of pre-trained UQ modules, we also conducted evaluations on Russian and Chinese prompts from \cite{vashurin2024benchmarking}, and additionally created a similar test set with German prompts. Test sets for each language consist of 100 biography-related questions. The statistics about all test sets are presented in \Cref{tab:test_stats}.

\paragraph{Metrics.} In the main experiments, we measured the claim-level performance of detecting invalid claims. For this purpose, we used PR-AUC, where ``unsupported'' claims represent the positive class.

\paragraph{Models.}
We conducted our primary experiments with Mistral 7b Instruct v0.2~\cite{mistral} and Gemma 2 9b Instruct \cite{team2023gemini}.

\paragraph{Training the uncertainty heads and hyper-parameter optimization.}

We trained the uncertainty heads using Adam with a linear learning rate decay and warmup. We selected the values of the hyper-parameters on the validation set of the \emph{biographies} dataset using claim-level PR-AUC metric and the Bayesian optimization algorithm available in the W\&B framework. We observed that among the important general hyper-parameters are the weight of the instances with positive labels, the number of epochs, and the size of the learning rate. The best values of the hyper-parameters for each of the tested models are presented in \Cref{tab:hyperparameters} in \Cref{sec:hyp}.

\paragraph{Baselines.} We compare our method to several unsupervised baselines: Maximum Claim Probability (an adaptation of Maximum Sequence Probability for claims), Mean Token Entropy, Perplexity, and Claim Conditioned Probability (CCP) \cite{fadeeva2024fact}. Additionally, we evaluated against supervised methods, including SAPLMA, Factoscope, and Lookback lens. SAPLMA predicts token-level uncertainties using a 3-layer perceptron, and the mean uncertainty is calculated over claim-related tokens during inference. Note that both Lookback lens and Factoscope operate at the claim level. Lookback lens uses a Logistic Regression model trained on attention features. Our implementation of the Factoscope approach uses our Transformer-based architecture and the feature set that includes hidden states, top token embeddings with similarities, and token ranks. The values of the hyper-parameters for the baselines selected after tuning are given in \Cref{sec:hyp}.

\subsection{Results}

\paragraph{Main results.}
Table~\ref{tab:results_train_bio_mistral} shows the performance of the unsupervised UQ techniques and the supervised UQ methods
trained on persons' biographies for claim-level hallucination detection with Mistral 7B Instruct v0.2. For evaluating supervised methods, the domain \emph{biographies} represents the in-domain test set and all other domains
represent out-of-domain (OOD) test sets. Note that in this evaluation, both the questions and the LLM's responses across all domains are in English.

Among the unsupervised techniques, uncertainty scores based on CCP yield the best performance, confidently outperforming other methods on \emph{biographies}, \emph{cities}, \emph{artworks}, and \emph{landmarks}. Mean Token Entropy also achieves relatively good results on par with CCP on \emph{books}, \emph{inventions}, and \emph{events}.

Supervised UQ methods greatly outperform unsupervised techniques on the in-domain test set. Moreover, remarkably, all considered supervised methods demonstrate substantial generalization and the ability to perform well beyond the training domain of people's biographies.

Our uncertainty head (UHead) demonstrates the best results in both in-domain and out-of-domain evaluations. For in-domain evaluation, UHead outperforms the best-unsupervised method CCP by 16 percentage points (pps) in terms of PR-AUC. The gap is also large for out-of-domain evaluation, e.g., for \emph{books}, UHead outperforms CCP by 23 pps, for \emph{movies} and \emph{events} by 20 pps, for \emph{artworks} by 18 pps, and for \emph{cities} by 12 pps.
Compared to supervised methods, UHead surpasses the closest competitor, Factoscope, by five pps for the in-domain evaluation. In OOD evaluation, it confidently outperforms other supervised methods across all domains, except for \textit{landmarks}, where it is slightly below the closest competitor by 1 pp.

Analyzing other supervised methods, the second-best scores are usually demonstrated by Factoscope.
We assume that the underperformance of the baseline based on the Factoscope features compared to UHead lies in the use of layer activations, which limits its generalization.
Another module that relies on hidden states is SAPLMA. In addition to the feature limitations, it also has architectural limitations, which further hurt its performance. For \emph{landmarks}, SAPLMA shows good results, but for other test sets, it stays behind Factoscope and UHead.
Compared to UHead, it lags by 12 percentage points on in-domain evaluation and up to 20 percentage points on out-of-domain evaluation.
Lookback lens also usually falls behind UHead and Factoscope; we believe that its main problem is its weak linear architecture.
At the same time, we note that the feature set suggested by Lookback lens based on attention is quite strong (see analysis of various feature sets below).

\paragraph{Introducing more diverse training data for UHead.}
Table \ref{tab:results_moredata} presents the results when we train uncertainty heads on \emph{biographies} plus the data from all domains except one, which is used for OOD evaluation. In this scenario, uncertainty heads get access to bigger and more diverse training data. As we can see, expanding the dataset provides slight improvements for certain domains. These results indicate that expanding the training data and enhancing its diversity could further increase the UQ performance, particularly in the OOD setting.

\paragraph{Analysis of feature sets.} Table \ref{tab:results_train_bio_features} presents the comparison of various feature sets in combination with the UHead architecture on the in-domain validation set. For each feature set, we perform an extensive hyper-parameter value search in the same way as for the main results.
We can see that all feature sets that leverage hidden states fall substantially behind attention-based features. The analysis of the validation loss dynamics shows that this is probably due to quick overfitting. Models that leverage hidden states start overfitting after 1--3 epochs, while models that leverage attention might not overfit even after 10 epochs. We also note that Lookback lens features combined with the UHead architecture provide strong performance. However, simple attention maps without feature engineering used in UHead yield even better results.

\paragraph{Cross-lingual generalization.}
Table \ref{tab:results_multilang} presents the results for Gemma 2 9b Instruct. In this experiment, we train UQ modules on the English person's \emph{biographies} as in the previous experiment, but we evaluate the performance on other languages. Surprisingly, UHead achieves strong cross-lingual generalization. For all OOD languages, UHead achieves substantial improvements over the best unsupervised methods. For Chinese, UHead is better than MTE by 10 pps; for Russian, it is better than CCP by 9 pps; and for German by 13 pps. Notably, other supervised methods also demonstrate some level of generalization, but in most cases, they have substantially worse performance. Overall, these results show that uncertainty heads, even if they are pre-trained on English data, can be good off-the-shelf hallucination detectors for LLM outputs in other languages.

\paragraph{Analysis of attention-based features.}

\begin{figure}[t]
  \centering
  \begin{subfigure}[b]{\linewidth}
    \centering
    \includegraphics[trim=0 0 0 0,clip,width=\linewidth]{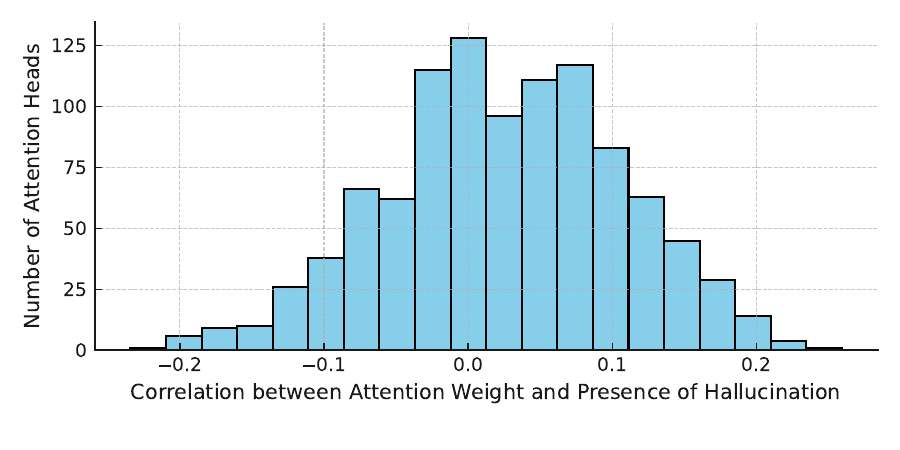}
    \caption{Correlation across various attention heads.}
    \label{fig:correlation_distribution}
  \end{subfigure}
  \vspace{1em}
  \begin{subfigure}[b]{\linewidth}
    \centering
    \includegraphics[trim=0 0 0 0,clip,width=\linewidth]{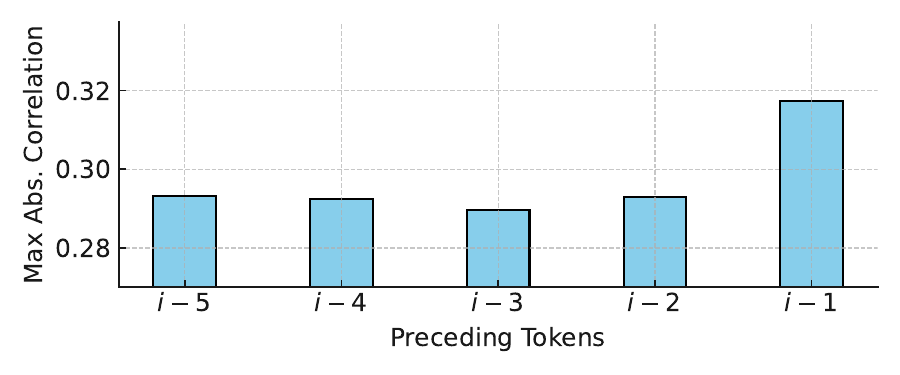}
    \caption{Max absolute attention–hallucination correlation.}
    \label{fig:attn_correlation}
  \end{subfigure}
  \caption{
    (a) The distribution of correlations between attention on the \(i-1\)-th token and presence of the \(i\)-th token in hallucinated claim. (b) The maximum absolute correlation across heads and layers for the same phenomenon. All scores were computed using the Mistral model and the \textit{biographies} dataset.
  }
  \label{fig:combined_correlations}
\end{figure}

In this analysis, we investigate attention patterns that may signal hallucinations in the currently generated token. Figure~\ref{fig:combined_correlations}a presents the correlation between attention weights from the generated token to the immediately preceding token and the presence of hallucination across various attention heads. While most heads show negligible correlation, a subset of heads exhibits moderate positive or negative associations. Figure~\ref{fig:combined_correlations}b further highlights that this correlation is strongest for the token immediately preceding the generated one. These findings suggest that a small number of attention heads encodes informative signals related to hallucination detection and reflect distinct model behavior under uncertainty during generation.

These findings are also confirmed by Figures \ref{fig:an1} and \ref{fig:an2}. Figure \ref{fig:an2} illustrates that attention weights from individual middle layers could serve as relatively strong hallucination detectors. Figure \ref{fig:an1} shows that
optimal performance is obtained by UHead when using attention weights from only 1--5 preceding tokens.

\paragraph{Computational efficiency.}

Next, we evaluated the computational overhead of various UQ methods.
To ensure a fair comparison, we focused only on the time required to generate texts and to compute uncertainty scores, excluding the time spent on claim extraction. Claim extraction could be performed by a small model specifically fine-tuned for this task, and its overhead is negligible compared to LLM inference. The results were obtained using a multi-domain dataset containing 800 texts and a total of 18,852 claims and  Mistral 7B Instruct v0.2.

Table \ref{tab:comp_efficiency} summarizes the results and provides the memory footprint of various methods. MCP and Perplexity incur no additional overhead, serving as baselines for comparison. The proposed UHead method introduces only 5\% overhead, which is slightly better than the best-unsupervised method CCP (8.6\%). With around 10 million parameters,
UHead has a minimal impact on the GPU memory footprint (40 MB). Thus, UHead is a very lightweight addition to multi-billion-parameter LLMs and is practical for real-world deployment.

\begin{figure}[t]
  \begin{adjustbox}{width=0.48\textwidth}
    \begin{minipage}{\linewidth}
      \begin{lstlisting}[language=Python]
from transformers import AutoModelForCausalLM, AutoTokenizer
from luh import AutoUncertaintyHead, CausalLMWithUncertainty

llm = AutoModelForCausalLM.from_pretrained(
    model_name)
tokenizer = AutoTokenizer.from_pretrained(
    model_name)
uhead = AutoUncertaintyHead.from_pretrained(
    uhead_name, base_model=llm)
llm_adapter = CausalLMWithUncertainty(llm, uhead, tokenizer=tokenizer)

# tokenize text and prepare inputs ...
output = llm_adapter.generate(inputs)
      \end{lstlisting}
    \end{minipage}
  \end{adjustbox}
  \caption{Code example for using uncertainty heads.}
  \label{code:example}
\end{figure}

\section{Collection of Pre-trained Uncertainty Heads for Popular LLMs}

Finally, we pre-trained a collection of Uncertainty Quantification (UQ) Heads for a range of popular 7B–9B parameter LLMs, including Mistral, various versions of LLaMA, and Gemma 2. In addition to model-level UQ, we release token-level UQ heads that can provide uncertainty scores directly for tokens without explicit claim annotation, which enables broader applicability across tasks.

Our UQ heads are designed for use as an off-the-shelf tool for confidence estimation in LLMs. They could be loaded from the hub using a procedure that is similar to the ``from\_pretrained'' API in the Hugging Face Transformers library and integrated into the LLM generation procedure with an adapter. A code example demonstrating how to use the UQ heads is provided in Figure \ref{code:example}. Thus, UQ heads could be integrated into third-party code with minimal modifications, which makes them an easy plug-and-play solution for researchers and practitioners.

\section{Conclusion and Future Work}

We presented pre-trained UQ heads -- supplementary supervised modules for LLMs that help to capture their uncertainty much more effectively than unsupervised UQ methods. We demonstrated that they are quite robust and deliver state-of-the-art results for both in-domain and out-of-domain prompts. They also show remarkable generalization to other languages.
Inspired by their good performance, we pre-trained a collection of UQ heads for a series of popular LLMs, including Mistral, Gemma 2, and LLama. We release the code and the pre-trained uncertainty heads so they could be used as off-the-shelf hallucination detectors for other researchers and practitioners.

We see that the performance of UQ heads improves with providing more training data from diverse domains. In future work, we plan to scale up the training data and explore the limits of the supervised approach to UQ.

\section*{Limitations}
Uncertainty heads cannot solve the problem when LLMs are trained to provide misinformation. In this situation, models are confident in their deceptive answers. Uncertainty heads cannot provide ideal annotation of hallucinations, as some LLMs do not have enough capacity to provide information about what they know and what they do not know. While we see generalization in uncertainty heads, we should acknowledge that, as with any other supervised method, they work best for ``in-domain'' data.
The bias present in LLMs could also be transferred into uncertainty heads.

\section*{Ethical Considerations}

\paragraph{Responsible Use}
  In our work, we considered open-weight LLMs and datasets not aimed at harmful content. However, LLMs may generate potentially damaging texts for various groups of people. Uncertainty quantification techniques can help create a more reliable use of neural networks. Moreover, they can be applied to detecting harmful generations, but this is not our intention.

\paragraph{Limited Applicability}
  Moreover, despite that our proposed method demonstrates sizable performance improvements, it can still mistakenly highlight correct and not dangerous generated text with high uncertainty in some cases. Thus, as with other uncertainty quantification methods, it has limited applicability.

\paragraph{Annotation Considerations}
We used GPT-4o for claim extraction and their annotation. This may introduce cultural, linguistic, or other biases into the data used to train the uncertainty heads.

\bibliography{custom}

\appendix

\clearpage
\onecolumn

\section{Training Data Generation Pipeline}
\label{sec:appendix}

\begin{figure*}[h]
    \centering
    \resizebox{1.\textwidth}{!}{
    \includegraphics[trim={0.3cm 3.cm 0.3cm 4.cm},clip,width=1.\linewidth]{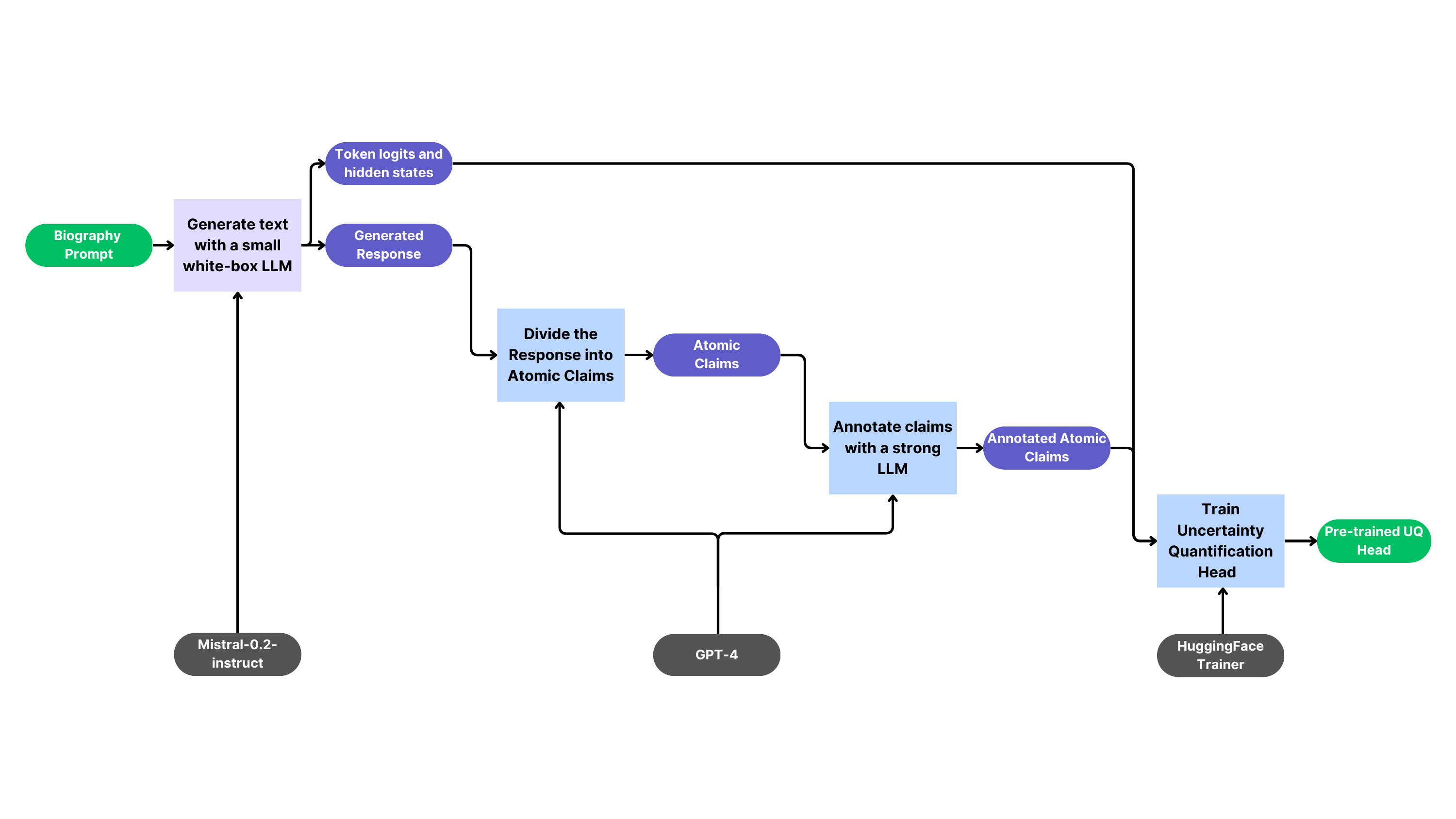}}
    \caption{The training data generation pipeline.}
    \label{fig:data_generation_pipeline}
\end{figure*}

\section{Dataset Details}

\subsection{Dataset Construction}~\label{sec:app-dataset-constr}

We used few-shot learning to better guide the LLM to generate the items for the desired domain. The structure of the prompts looks as follows:

\begin{lstlisting}[style=mdstyle]
Continue the list of 100 most famous {domain items}:

1. <domain-item-1>

2. <domain-item-2>

3. <domain-item-3>
\end{lstlisting}

\vspace{0.7cm}
Example for the ``cities'' domain:

\begin{lstlisting}[style=mdstyle]
Continue the list of 100 most famous cities:

1. Paris, France

2. Amsterdam, Netherlands

3. Osaka, Japan
\end{lstlisting}
\vspace{0.7cm}

For claim extraction and their annotation, we use GPT-4o with prompts from \cite{fadeeva2024fact}. Overall expenses for LLM API calls are approximately \$4000.

\vspace{3.5cm}

\subsection{Dataset Statistics}
\label{sec:app-dataset-stats}

Table \ref{tab:train_stats} presents the statistics of the datasets used for training and validation; Table \ref{tab:test_stats} shows the statistics of the datasets used for testing.

\begin{table}[h]

\centering
\footnotesize

\resizebox{0.5\textwidth}{!}{

\begin{tabular}{l|l|c|c}
\toprule
\textbf{Model} & \textbf{Dataset} & \textbf{\#} \textbf{of texts} & \textbf{\# of claims} \\
\midrule
\multirow{2}{*}{Mistral 7b Instruct v0.2} & biographies & 3,300 & 57,671 \\
 & multi-domain & 700 & 14,554 \\
 \midrule
\multirow{1}{*}{Gemma 2 9b Instruct} & biographies & 3,300 & 83,716 \\
\bottomrule
\end{tabular}

}

\caption{Statistics about the training datasets used in our experiments.}
\label{tab:train_stats}
\end{table}

\begin{CJK}{UTF8}{gbsn}
\begin{table*}[h]

\centering
\footnotesize
\resizebox{\textwidth}{!}{

\begin{tabular}{l|c|l|l|cc|cc}
\toprule
\textbf{Split} & \textbf{\# of prompts} & \textbf{ChatGPT prompt used to generate questions} & \textbf{Testing prompt} & \multicolumn{2}{c}{\textbf{\# of claims}} & \multicolumn{2}{c}{\textbf{Accuracy}} \\
 & & & & \textbf{Mistral} & \textbf{Gemma} & \textbf{Mistral} & \textbf{Gemma} \\
\midrule
persons & 100 & Tell me a list of 100 most famous persons. & Tell me a bio of a <person> & 2234 & 2857 & 72.9\% & 87.4\% \\
cities & 100 & Tell me a list of 100 most famous cities. & Tell me a history of a <city> & 2128 & 2684 & 79.8\% & 87.1\% \\
movies & 100 & Tell me a list of 100 most famous movies.  & Tell me about the movie <movie> and its cast. & 2568 & 3121 & 89.7\% & 94.8\% \\
inventions & 100 & Tell me a list of 100 most important inventions. & Tell me about the invention of <invention> and its inventor. & 2269 & 2626 & 84.3\% & 92.1\% \\
books & 100 & Tell me a list of 100 most famous books. & Tell me about the book <book> and its author. & 2530 & 3070 & 89.9\% & 95.9\%  \\
artworks & 100 & Tell me a list of 100 most famous artworks. & Tell me about the artwork <artwork> and its artist. & 2464 & 2873 & 75.9\% & 85.1\%  \\
landmarks & 100 & Tell me a list of 100 most famous landmarks. & Tell me about the landmark <landmark>. & 2365 & 2566 & 88.5\% & 93.7\% \\
events & 100 & Tell me a list of 100 most significant historical events. & Tell me about <event> event. & 2294 & 2665 & 88.9\% & 94.8\% \\
\midrule
Russian & 100 & --- & Расскажи биографию <person> & --- & 3572 & --- & 66.7\%  \\
Chinese & 100 & --- & 介绍一下<person> & --- & 2248 & --- & 77.8\%  \\
German & 100 & --- & Erzählen Sie mir eine Biografie von <person> & --- & 2815 & --- & 85.1\% \\
\bottomrule
\end{tabular}

}

\caption{The statistics of the multi-domain test dataset and number of claims generated my Mistral 7B Instruct v0.2 and Gemma 2 9b Instruct models.}
\label{tab:test_stats}
\end{table*}
\end{CJK}

\section{Hyperparameters}
\label{sec:hyp}







\begin{table}[h]
    \resizebox{\textwidth}{!}{
    \centering
    \begin{tabular}{l|l|c|c|c|c|c|c|c}
    \toprule
    \textbf{Method} & \textbf{Model}  & \textbf{Learning Rate} & \textbf{Num. Epochs} & \textbf{Weight Decay} & \textbf{Dropout Rate} & \textbf{Warmup} & \textbf{Att. Window Size} & \textbf{Transformer arch.} \\
    \midrule
    \multirow{2}{*}{SAPLMA}
    & Gemma 2 9b Instruct & 1e-4 & 10 & 0.1 & 0.1 &  & -- & --  \\
    & Mistral 7b Instruct v0.2 & 1e-4 & 10 & 0.1 & 0.1 &  & -- & -- \\
    \midrule
    \multirow{2}{*}{Lookback lens}
    & Gemma 2 9b Instruct & 1e-2 & 13 & 0.1 & 0.1 &  & -- & -- \\
    & Mistral 7b Instruct v0.2 & 1e-2 & 13 & 0.1 & 0.1 &  & -- & -- \\
    \midrule
    \multirow{2}{*}{UHead (Factoscope)} 
    & Gemma 2 9b Instruct & 2e-4 & 3 & 0.1 & 0.2 &  & -- & \\
    & Mistral 7b Instruct v0.2 & 2e-4 & 5 & 0.2 & 0.2 & 0.05 & -- & 1 layers / 256 width / 4 heads \\
    \midrule
    \multirow{2}{*}{UHead} 
    & Gemma 2 9b Instruct & 2e-4 & 6 & 0.1 & 0.05 &  0.1 & 2 & 1 layer / 768 width / 16 heads \\
    & Mistral 7b Instruct v0.2 & 2e-4 & 7 & 0.1 & 0.2 & 0.1 & 2 & 2 layers / 768 width / 4 heads \\
    \bottomrule
\end{tabular}
    }
    \caption{Optimal hyperparameters for each method and model.}
    \label{tab:hyperparameters}
\end{table}

For each tested model, we selected hyperparameters by optimizing the PR-AUC metric on the validation set of the ``biographies'' dataset. In training, we optimized the learning rate, warmup ratio, number of epochs, and the weight of positive examples in the cross-entropy loss. For the model architecture, we optimized the number of uncertainty layers, the number of heads, and the intermediate dimension. For feature extraction, we optimized the number of layers used to obtain hidden states, token probabilities, and attention weights, as well as the number of preceding tokens considered for attention. The optimal hyperparameters are summarized in Table~\ref{tab:hyperparameters}.
The hyperparameter grid is the following:
\begin{itemize}[left=0pt,noitemsep,label={}]
    \item \textbf{Learning rate:} \{1e-5, 3e-5, 5e-5, 1e-4, 2e-4, 5e-4, 1e-2\};
    \item \textbf{Num. of epochs:} \{$n \in \mathbb{N} \mid 2 \leq n \leq 15$\};
    \item \textbf{Warmup:} \{0., 0.05, 0.1\};
    \item \textbf{Attention window size:} \{1, 2, 3, 4, 5, 10\};
    \item \textbf{Dropout rate:} \{0., 0.05, 0.1, 0.2\};
    \item \textbf{Weight decay:} \{0, 1e-2, 1e-1\}.
\end{itemize}

\clearpage
\section{Analysis of Attention-Based Features}

\begin{figure}[h]
    \centering
    \includegraphics[width=0.6\textwidth]{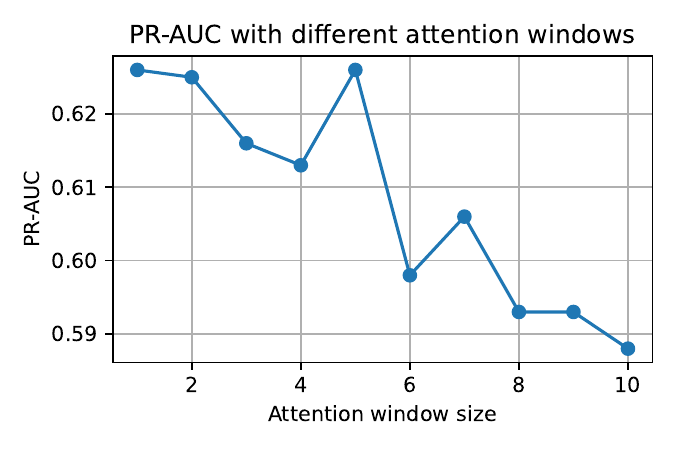}
    \caption{PR-AUC for different attention window sizes using UHead for Mistral 7B Instruct v0.2 model.}
    \label{fig:an1}
\end{figure}
\begin{figure}[h]
    \centering
    \includegraphics[width=0.7\textwidth]{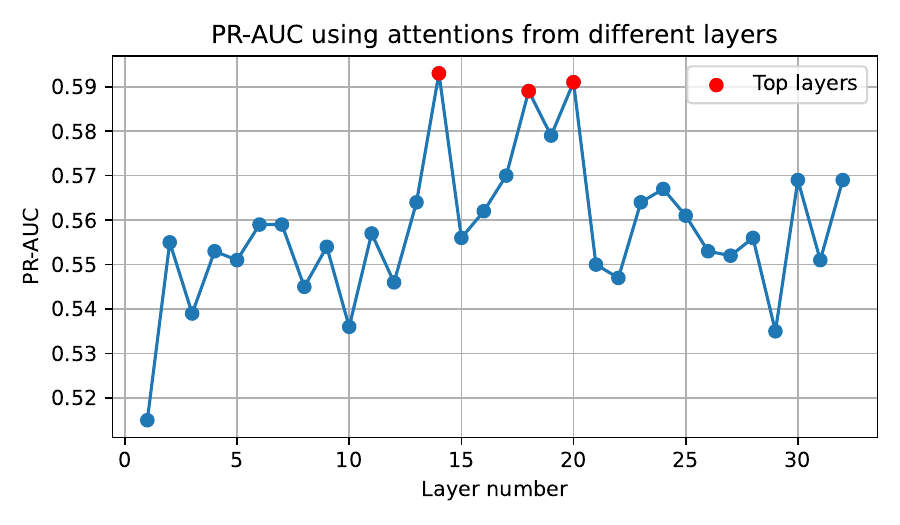}
    \caption{PR-AUC as a function of layer number used for attention features in UHead for Mistral 7B Instruct v0.2 model. Highlighted points mark layers with highest PR-AUC (layers 14, 18 and 20).}
    \label{fig:an2}
\end{figure}

\clearpage
\section{Hardware and Computational Efficiency}

All experiments were conducted on 8 NVIDIA RTX 5880 Ada GPUs. On average, training a single model with hyperparameter search takes around 150 GPU hours.

\begin{table}[ht]

\centering
\footnotesize


\begin{tabular}{l|c|c}
\toprule
\textbf{Method} & \textbf{Computational Overhead} & \textbf{GPU Memory Footprint} \\
\midrule
MCP & 0.0 \% & - \\
Perplexity & 0.0 \%  & - \\
Max Token Entropy & 0.2 \% & - \\
CCP & 8.6 \% & 1,546 MB \\
SAPLMA & 4.7 \% & 4 MB \\
Factoscope & 6.1 \% & 32 MB \\
Lookback Lens & 5.5 \% & <1 MB \\
UHead (only hidden states) &  & 73 MB  \\
UHead (att. + prob. + hs.) &  & 82 MB \\
UHead (Ours) & 4.9 \% & 40 MB \\ 
\bottomrule
\end{tabular}


\caption{Computational overhead of UQ methods evaluated with the Mistral 7B Instruct v0.2 model. Overhead is measured relative to the fastest method, MCP. For CCP, the size of the auxiliary NLI model is reported.}
\label{tab:comp_efficiency}
\end{table}

\end{document}